\documentclass[tablecaption=bottom,wcp]{jmlr} % W&CP article
\usepackage[utf8]{inputenc}

\usepackage{booktabs}

\theorembodyfont{\upshape}
\theoremheaderfont{\scshape}
\theorempostheader{:}
\theoremsep{\newline}

\usepackage{caption}
\usepackage{subcaption}
\usepackage{listings, chngcntr, float, amsmath, amssymb, cmap, graphicx, xcolor, hyperref}
\usepackage{amsmath, multirow}

\usepackage{color}
\usepackage{grffile}
\usepackage{array,tabulary}
\usepackage{wrapfig}
\graphicspath{{media/}}

\usepackage{mathtools,mathptmx}
\mathtoolsset{showonlyrefs=true}

\usepackage[mathcal]{euscript}
\usepackage{amsfonts}
\usepackage{tikz}
\usetikzlibrary{arrows.meta,shapes.misc}

\jmlrvolume{60}
\jmlryear{2018}
\jmlrworkshop{Conformal and Probabilistic Prediction and Applications}
\jmlrproceedings{PMLR}{Proceedings of Machine Learning Research}

\title[fMRI: preprocessing, classification and pattern recognition]{fMRI: preprocessing, classification and pattern recognition}

\newcommand*{\mathcolor}{}
\def\mathcolor#1#{\mathcoloraux{#1}}
\newcommand*{\mathcoloraux}[3]{%
  \protect\leavevmode
  \begingroup
    \color#1{#2}#3%
  \endgroup
}

\author{\Name{Maxim Sharaev} \Email{m.sharaev@skoltech.ru}\\
\addr Skolkovo Institute of Science and Technology,  Skolkovo, Moscow Region, Russia
\AND
\Name{Alexander Andreev} \Email{aandreev@deloitte.ru}\\
\addr Deloitte Analytics Institute, Moscow, Russia
\AND
\Name{Alexey Artemov} \Email{a.artemov@skoltech.ru}\\
\addr Skolkovo Institute of Science and Technology,  Skolkovo, Moscow Region, Russia
\AND
\Name{Alexander Bernstein} \Email{a.bernstein@skoltech.ru}\\
\addr Skolkovo Institute of Science and Technology,  Skolkovo, Moscow Region, Russia
\AND
\Name{Evgeny Burnaev} \Email{e.burnaev@skoltech.ru}\\
\addr Skolkovo Institute of Science and Technology,  Skolkovo, Moscow Region, Russia
\AND
\Name{Ekaterina Kondratyeva}\Email{ekaterina.kondratyeva@skolkovotech.ru}\\
\addr Skolkovo Institute of Science and Technology,  Skolkovo, Moscow Region, Russia
\AND
\Name{Svetlana Sushchinskaya} \Email{svetlana.sushchinskaya@skolkovotech.ru}\\
\addr Skolkovo Institute of Science and Technology,  Skolkovo, Moscow Region, Russia
\AND
\Name{Renat Akzhigitov} \Email{barms@yandex.ru}\\
\addr Moscow Research and Clinical Center for Neuropsychiatry, Moscow, Russia\\
Skolkovo Institute of Science and Technology,  Skolkovo, Moscow Region, Russia}

\editors{Alex Gammerman, Vladimir Vovk, Zhiyuan Luo, and Harris Papadopoulos}% for multiple editors

\begin{document}

\maketitle
\medskip

\begin{abstract}
  As machine learning continues to gain momentum in the neuroscience community, we witness the emergence of novel applications such as diagnostics, characterization, and treatment outcome prediction for psychiatric and neurological disorders, for instance, epilepsy and depression. Systematic research into these mental disorders increasingly involves drawing clinical conclusions on the basis of data-driven approaches; to this end, structural and functional neuroimaging serve as key source modalities. Identification of informative neuroimaging markers requires establishing a comprehensive preparation pipeline for data which may be severely corrupted by artifactual signal fluctuations. In this work, we review a large body of literature to provide ample evidence for the advantages of pattern recognition approaches in clinical applications, overview advanced graph-based pattern recognition approaches, and propose a noise-aware neuroimaging data processing pipeline. To demonstrate the effectiveness of our approach, we provide results from a pilot study, which show a significant improvement in classification accuracy, indicating a promising research direction.
\end{abstract}


\section{ Introduction}

  One challenge for the successful applications of automated diagnostics based on pattern recognition approaches in clinical settings is to ensure the highest possible quality of source signal employed for decision making. Cleaning the artifactual (irrelevant to the process of question) noise incidental to scanning deems necessary, as such fluctuations drastically hurt recognition performance, blocking the way to the identification of neuroimaging markers for mental disorders~\cite{birn2009fmri}. To this end, denoising schemes must be proposed, which involve the extensive examination of spatiotemporal constituents of the source signal and identification of the relevant components against the artifactual noise~\cite{bianciardi2009sources,salimi2014automatic}. In the present work, we investigate a pattern classification pipeline for mental disorders featuring a denoising step, and observe consistent performance improvements w.r.t. the baseline approach.

A different but important challenge is to design highly sensitive and robust predictive models. Research indicates that the pattern of brain activity changes associated with disorders such as depression might have limited discriminative power, leading to performance drops for common machine learning algorithms. Current accuracy of around 75\%, thus, does not allow direct clinical application of these models~\cite{kambeitz2015detecting}. However, recent advancements on graph theory such as persistent homology-based analysis and its applications in neuroimaging indicate that features extracted from connectome should be informative and robust enough markers for pattern classification~\cite{stolz2017persistent}. Effective connectivity and topological data analysis thus represent special interest for decision making in clinical applications. In this paper, we only investigate a basic graph-based features that already prove to be effective in discriminating between two groups of patients: only epilepsy and both epilepsy$+$depression patients \cite{BurnaevBernstein}. We note that application of advanced methods poses denoising as a preliminary step, as noisy data is not reliable enough to build a connectome. Thus, both steps are essential.

  In summary, our contributions in this work are the following:
  \begin{itemize}
\item We systematically review common approaches to preparation, preprocessing and denoising of neuroimaging data aimed at the elimination of artifacts harmful for pattern recognition;
\item We systematically review advanced approaches to data representation for pattern recognition, with much attention devoted to graph-based methods and topological data analysis as the promising areas;
\item We propose a principled noise-aware pattern recognition pipeline for neuroimaging tailored to pattern classification;
\item Using a pilot study that aims to classify between patients with epilepsy only and patients with both epilepsy + depression, we demonstrate the effectiveness of our proposed methodology.
\end{itemize}

This paper is structured as follows. In section 2, we make an affordable introduction to functional neuroimaging and review its applications in the field of clinical neurology. Section 3 describes in detail the possible sources of fluctuations arising in functional neuroimaging, as suggests how these fluctuations can be identified by the means of data analysis. In section 4, we briefly review approaches to inference and analysis of the connectome via the techniques of graph-based analysis. Section 5 presents our proposed noise-aware data processing pipeline and the results of the pilot study, that highlight the advantages of our approach.

\section{Functional MRI and its applications in neurology}

\subsection{Neuroimaging and fMRI}

  Magnetic resonance imaging (MR imaging, MRI) and its part - functional magnetic resonance imaging (fMRI) - use strong magnetic fields to create images of biological tissues and physiological changes in them. The static magnetic field created by an MRI scanner is expressed in units of Tesla (T). Commercial MR scanners are usually within the range of 1.5T to 3T, whereas research-oriented scanners could have stronger fields of 7T and more. To create images, the scanner uses a series of changing magnetic gradients and oscillating electromagnetic fields, known as a pulse sequence. Depending on the resonance frequency, energy from the electromagnetic fields is absorbed by the atomic nuclei. MRI scanners operate at the frequency of hydrogen nuclei, the most common particle in the human body due to their prevalence in water molecules~\cite{huettel2004functional}. 

Depending on the pulse sequence, it is possible to create static images of tumors, blood vessel damages, and other tissue abnormalities. As of 2018, it is possible to obtain individual MR slices every 50-100ms (an acquisition mode known as echo-planar imaging or EPI), thus changes of tissue properties in time could be captured, producing a dynamic picture of the brain. Functional neuroimaging attempts to assign different mental processes to different parts of the brain, creating a “brain map” of areas responsible for respective cognitive processes~\cite{huettel2004functional}. Functional MRI is a technique to measure changes in blood oxygenation over time. Blood oxygenation level is highly correlated with the activity of neurons~\cite{logothetis2001neurophysiological} in a brain region, so it becomes possible to measure their activity (though indirectly), obtaining the temporal resolution within seconds and the spatial resolution within up to 1mm or even better in high-field scanners. The sampling rate for fMRI varies from a few hundred milliseconds to a few seconds, which is much faster than in positron-emission tomography (PET) studies where a single image could take as long as one minute to capture.
      
  A number of manufacturers of MR scanners exist; the most significant are Siemens, Philips, and General Electric. Though there are minor differences in hardware and software parts, main components of the scanner are always the same. The static magnetic field produced by the main superconductive magnet is an absolute necessity for MRI. The two main criteria for characterizing a static magnetic field are its homogeneity and strength~\cite{huettel2004functional}. In case of inhomogeneity, the signal measured from a volumetric pixel or “voxel” would depend upon its location in the magnetic field. The magnitude of nuclear resonance depends on the field strength, thus making the latter crucial for obtaining greater values of the signal-to-noise ratio (SNR).

MR signal is produced by transmitter and receiver radiofrequency (RF) coils that generate and receive electromagnetic pulses at the resonant frequency of the specific nuclei of the studied tissue. Now RF coils are constructed with an increased number of individual elements, showing improvements in several characteristics of MRI quality. A 32-channel array head coil is reported to have exhibited substantial gains in SNR compared to an 8-channel coil~\cite{wiggins200996} and an overall higher SNR and reduced noise amplification for accelerated imaging compared to a 12-channel coil at 3T~\cite{wiggins200996}. It was shown that the 32-channel coil detected more fMRI-activation cortically but less subcortically than the 12-channel coil~\cite{kaza2011comparison}. Gradient coils are another crucial part of a scanner. Gradient coils are used to spatially encode the positions of voxels by varying the magnetic field linearly across the imaging volume. The resonant frequency will then vary as a function of voxel position in space. Gradient coils are powered by sophisticated amplifiers allowing rapid and precise adjustments to field strength and direction~\cite{hidalgo2010theory}. Typical gradient systems are capable of producing gradients from 20–100 mT/m. The gradient system performance influences scan speed: stronger gradients allow for faster imaging or for higher resolution; gradient systems capable of faster switching can also permit faster scanning~\cite{gambhir2010molecular}. The slew rate of a gradient system is a measure of how quickly the gradients can be switched on or off. Nowadays gradients have a slew rate of up to 100–200 $\mathrm{T}\cdot \mathrm{m}^{-1} \cdot \mathrm{s}^{-1}$. The slew rate significantly impacts the image quality and depends both on the size of the gradient coil (bigger coils are slower) and on the performance of the amplifier (it must provide high voltage because of the inductance of the coil).

\subsection{Univariate analysis of fMRI data}

Since the introduction of fMRI, the central research design was to compare the measurements at each voxel statistically using T-tests which revealed the change in each voxel’s activity depending on the experimental design and was used to construct a “brain map”, indicating the distribution of neural correlates related to a particular phenomenon~\cite{friston1994statistical}. This simplistic approach has some limitations. First, it is dealing with binary comparisons, which can be underpowered when identifying continuous processes~\cite{cohen2017computational}. To correct for binary comparisons, parametric designs were introduced, which used regression to identify voxels responsive in a predicted way~\cite{braver1997parametric}. Second, the volume and extraordinary dimensionality of brain imaging data result in a large number of statistical comparisons leading to an increased risk of false discoveries; in response, methods of correcting for multiple comparisons were developed~\cite{forman1995improved,smith2009threshold}. Presently, there is a consensus about the most important kinds of preprocessing (slice-time correction, motion correction, filtering in space and time, anatomical alignment across individuals), which resulted in widespread adoption of standard software toolboxes and pipelines of data preprocessing~\cite{cohen2017computational}. This unification and standardization foster reproducibility tests and comparison between studies, which becomes very important with the development of international databases and novel mathematical, statistical and computational techniques.

\subsection{Multivariate analysis and its applications}

As mentioned before, the first generation fMRI analysis methods were univariate or mass-univariate - they examined individual voxels or groups of voxels independently of each other. One attempt to overcome the limitations of univariate analysis is multi-voxel pattern analysis (MVPA). MVPA examines spatial patterns of activity over possibly distinct groups of voxels, to recover what information they represent collectively~\cite{lewis2013multi}. Two principal strategies of MVPA are widely employed in neuroimaging. The first approach is to apply classifiers from machine learning, mainly linear models, to discriminate between states (different types of processing, diagnosis and etc.) One challenge for this general approach is that classifiers readily exploit any discriminative variability, making it essential to analyze all possible variables that could lead to differences in groups~\cite{cohen2017computational}, such as task difficulty, reaction time in task-fMRI~\cite{todd2013confounds} and patient movements or tremor in resting-state fMRI.

  The second approach of MVPA deals with similarity/discrepancy measures voxel between patterns in a high-dimensional voxel space~\cite{cohen2017computational}. An example is Representational Similarity Analysis (RSA), where the representation in each brain region is characterized by a representational dissimilarity matrix, which is a square symmetric matrix, each entry referring to the dissimilarity between the activity patterns associated with two stimuli or experimental conditions~\cite{kriegeskorte2008representational}. 
    
  Applications of multivariate analysis of fMRI data and in particular machine learning seem very promising in the medical field. The number of clinical studies where task and resting-state fMRI is used in classification (diagnostics or a prognosis of treatment) and regression problems (presurgical planning, epilepsy foci localization) is rapidly growing. We demonstrate some recent examples of successful medical applications in sections~\ref{sec:schiza}--\ref{sec:epilepsy}.

\subsubsection{Schizophrenia diagnostics/treatment}
\label{sec:schiza}

  Schizophrenia is often associated with abnormal brain connectivity. The abnormality is referred to as a “dysconnection” hypothesis~\cite{stephan2009dysconnection} where “dysconnection” means a range of network dysfunctions, as opposed to “disconnection”, or reduced connectivity~\cite{gheiratmand2017learning}. Any biomarker of schizophrenia needs to be stable in terms of its predictive power on previously unseen subjects, which is usually not the case in schizophrenia due to the high variability of symptoms and illness severity. \cite{gheiratmand2017learning} investigated whole-brain fMRI functional networks, both at the voxel level and lower-resolution super-voxel level. Their target was Auditory Oddball task data on the FBIRN fMRI dataset ($n = 95$ subjects), where node-degree and link-weight network features were considered and generalization accuracy of statistically significant feature sets in discriminating patients vs. controls was evaluated. Sparse multivariate regression (elastic net) to whole-brain functional connectivity features helped derive stable predictive features for symptom severity. Whole-brain link-weight features reached 74\% accuracy in identifying deceased subjects and were more stable compared to voxel-wise node-degrees.  Link-weight features predicted the severity of several symptoms, including inattentiveness and bizarre behavior. The most significant, stable and descriptive functional connectivity features included increased correlations between the thalamus and the primary motor/primary sensory cortex, and between the precuneus (BA7) and the thalamus, the putamen, and the Brodmann areas BA9 and BA44~\cite{gheiratmand2017learning}. 

  Other mattering potential applications of fMRI and machine learning in schizophrenia include detection of fMRI signal changes during symptomatic events like hallucinations.  When applied to whole-brain fMRI data, conventional classification methods, such as the support vector machines (SVM), provide solutions that are difficult to interpret. In a recent work, it was proposed to extend the existing sparse classification methods by taking the spatial structure of brain images into account with structured sparsity using the total variation penalty~\cite{de2018prediction}. A reliable classification performance and interpretable predictive spatial patterns consisting of two clusters in speech-related brain regions were obtained. The characterization of the variability within pre-hallucination patterns (using an extension of principal component analysis with spatial constraints) helped to shed light on the intrinsic sources of the variability present in the dataset. These results could be promising for fMRI-guided therapy for drug-resistant hallucinations~\cite{de2018prediction}.

\subsubsection{Presurgical planning}
\label{sec:surgical}

  Brain lesions and diseases affect neural processing and increase individual variations in patients’ functional networks compared with healthy controls. It is of great importance to find functional brain areas before the surgery in order to maximize the resection area (in case of tumors) and minimize the influence on brain functions. A recent work investigates the ability to predict individual language areas from resting-state fMRI (rs-fMRI)~\cite{jones2017resting}. A predictive model supplied with pairs of resting-state and task-evoked data was trained to predict activation of unseen patients and healthy controls based on their resting-state data alone. Patients generally show higher variability compared to healthy controls in the location of language areas. However, the models were able to successfully learn variations in both patient and control responses from the individual resting-connectivity features. A kind of transfer learning was applied here: a model trained exclusively on the more-homogenous control group was shown to be able to predict task activations in patients~\cite{jones2017resting}. A serious limitation of this work is that the ground truth for resting-state fMRI prediction was task fMRI data, and not, for example, intraoperative mapping, which is nowadays a gold standard~\cite{lu2017automated}. This shortcoming was overcomed in another recent work, where resting-state fMRI has been also applied to pre-surgical functional mapping~\cite{lu2017automated}. The authors developed an automated method for identifying language network in brain tumor subjects using ICA on resting-state fMRI. In addition to standard processing strategies, a discriminability index-based component identification algorithm was applied to identify language networks in three different groups. The results obtained with the training group were validated in an independent group of healthy human subjects. For the testing group, the detected language networks were assessed by intraoperative stimulation mapping to verify the reliability of application in the clinical setting. On an individual level, language areas could be successfully mapped for all subjects from the two healthy groups except one (19/20, success rate = 95.0\%). In the testing group (brain tumor patients), the sensitivity of the language mapping result reached 60.9\% in case of exact match and 87.0\% in within the radius of 1 cm~\cite{lu2017automated}. 

\subsubsection{Depression diagnostics/treatment}
\label{sec:depression}

  For decades, neuroimaging has been widely used to explore the pathomechanism of major depression disorder (MDD), bipolar depression (BD) and related disorders. Some of the aspects covered in the literature are: identification of depressed vs. non-depressed subjects, differentiation between major (unipolar) depression and bipolar depression, prediction of the first onset of depression in young people (e.g. early onset vs. later onset), estimation of depression valence, assessment of effects of medication (e.g. antidepressants), prediction from resting-state fMRI (rs-fMRI), and selection of interpretable features.
\cite{patel2016studying} gives an overview of research into depression diagnostics and treatment response prediction using neuroimaging data and machine learning methods. As principled depression diagnostics probably remains the most important issue in literature, we review some of these works in greater detail. The development of features used by most studies has primarily focused on extracting information from T1-weighted imaging and fMRI. The use of these features in an initial attempt to model depression diagnosis and treatment response is in accordance with past neuroimaging-based studies that have predominantly found anatomical changes and altered brain activity to be valuable biomarkers of major depression~\cite{dunlop2014neuroimaging,mcgrath2013toward}. Most studies mentioned by \cite{patel2016studying} use linear support vector machines (SVMs) that offer reliable theoretic foundations and insensitivity to high-dimensional data, with other approaches including Gaussian processes, neural networks, random forests, k-means and decision trees~\cite{kambeitz2015detecting}. Nearly all approaches rely on leave-one-subject-out cross-validation (LOOCV) for characterization of the accuracy of the prediction models due to the scarcity of available data volumes.

\subsubsection{Epilepsy diagnostics/treatment}
\label{sec:epilepsy}

	Brain changes associated with temporal lobe epilepsy (TLE) include asymmetrical distribution of temporal lobe abnormalities on the same brain hemisphere in the hippocampus, parahippocampal gyrus, and entorhinal cortex~\cite{keller2008voxel}. Therefore many forthcoming studies attempted to find biomarkers of mTLE in these particular regions. For instance,~\cite{liao2011default} found reductions in both functional and structural connectivity between hippocampal structures and adjacent brain regions, as well as connectivity among the default mode network (DMN). It is believed that degeneration of structural connectivity may help explain the pathophysiological mechanism of the impaired cognitive functions. In voxel-level fMRI analysis, the basic experimental methodology involves defining regions of interest (ROIs) to be used as masks, applying them onto the residual images to extract the mean signal time-courses from each predefined ROI, and computing correlation coefficients between pairs of signal time-courses. The latter are then utilized to test hypotheses~\cite{bettus2010role}.

  Functional MRI-based research into TLE is typically performed with presented visual stimuli (words, faces or scenes), and the goal of the patient is to respond, for instance, by pushing the button~\cite{golby2002memory,bonnici2013assessing}. Assessing the functional reserve of key memory structures, which remains a challenge for pre-surgical patients with intractable temporal lobe epilepsy, is central for a number of studies~\cite{bonnici2013assessing}. They provide evidence for detection of predictable patterns of activity across voxels associated with specific memories in MTL structures, including the hippocampus. Overall, their findings indicate that MVPA-fMRI could prove a useful non-invasive method of assessing pre-surgical memory capacity within the MTL. Patients with long-standing epilepsy may have variable anatomic localization of neurologic functions, such as memory, because of cerebral reorganization induced by the disease process~\cite{golby2002memory}. Understanding this functional anatomy is vital when planning surgical resections and relies on complex preoperative evaluations. fMRI is a valid tool for assessing of memory lateralization in patients with MTLE and may therefore allow noninvasive preoperative evaluation of memory lateralization. Another issue is the development of neuroimaging measures that prove to be strongly predictive neuroimaging markers in pattern classification between healthy controls and general epileptic patients~\cite{zhang2012pattern}. Using modern pattern-recognition approaches like sparse regression and support vector machine, they have achieved a cross-validated classification accuracy of 83.9\% (specificity: 82.5\%; sensitivity: 85\%) across individuals from a large dataset consisting of 180 healthy controls and epileptic patients.

\section{Handling artifactual fluctuations in functional MRI data}

\subsection{fMRI-related noise modalities}

  Not depending on equipment, fMRI signal is very noisy. As T2*-weighted image (BOLD-contrast) is a mixture of signals from many sources, the desired signal from the neuronal activity only represents a relatively small percentage of the variance of the signal~\cite{bianciardi2009sources}. Non-neuronal contributions to the BOLD fMRI time series include receiver coil thermal noise, instrumental drifts, spike-like artifactual signals induced by the hardware instabilities, rapid and high-amplitude spikes due to the head motion. The physiological noise of non-neuronal origin (which is essentially BOLD-signal, but of no interest) comprises of cardiac and respiratory noise, changes in arterial carbon dioxide concentration associated with varying respiration rate, vasomotor effects, changes in blood pressure and cerebral autoregulation mechanisms~\cite{murphy2013resting}. We further discuss the major fluctuation sources and noise models. 

  As mentioned previously, fMRI data are indirect measurements of neural activity in a brain - the activity is proxied by BOLD (blood oxygen level dependent) response. In addition to meaningful BOLD signal, data contains noise component. This noise component is a combination of fluctuations with different origins and properties, below classification for each type of the fluctuations and possible approaches to reduce its influence are given~\cite{greve2013survey,welvaert2013simulation,caballero2017methods,liu2016noise}.

\emph{Background noise} includes thermal and system- or scanner-related fluctuations. \emph{Thermal noise} is caused by the Brownian motion of ions within the subject and the scanner. This motion leads to fluctuations in the electromagnetic field measured by the scanner. Such fluctuations always exist, but, in some sense, this noise is the easiest to model as its magnitude linearly depends on the temperature inside the scanner and the strength of its magnetic field, and it can be adequately modeled by spatially and temporally independent homoscedastic Gaussian noise. Its influence can be reduced either during the measurement session by decreasing the temperature in a scanner room and post-measurement via data averaging. \emph{System noise} comes from variations and instabilities in a fMRI scanner and results in a low-frequency drift. Though this type of noise is hard to assess, its influence is estimated to be a few percents from total variance~\cite{greve2011novel}, offering approaches to mitigate it. As follows from their definitions, both thermal and system noise exist independently from the subject and experiment and characterize the environment and the scanner, respectively.

\emph{Motion noise} comes from subject’s motion in a scanner.  This noise is almost inevitable and the effect of the noise is dramatic as it leads to voxels' movements. In task-based fMRI, head movements might correlate with experimental tasks such as, for instance, during speech-related tasks. While for the task-based fMRI effect of the noise might be reduced by averaging across trials, in resting-state fMRI even micro-movements might drastically affect the analysis. The effects of motion can be reduced by applying a motion correction algorithm~\cite{cox1999real}.

  \emph{Physiological noise} is primarily caused by the subject's heartbeat (or cardiac pulsatility) and air circulation (or respiration). Cardiac pulsatility cause artifacts related to brain tissue movements and blood vessels inflow effects. Respiration, in addition to small head movements (that might be related to motion noise), affects the carbon dioxide level in the blood that causes blood vessels to dilate and therefore affects the BOLD signal. Despite the relatively minor influence of the noise on the signal, these activities should also be taken into account: the heartbeat and respiratory rates could be measured but will require the involvement of additional external devices as its rate is greater than the Nyquist rate in most fMRI studies~\cite{greve2011novel}. 

\emph{Non-task-related noise} includes any non-task-related neural activity including, for instance, the reaction to the scanner acoustic noise or some subjects’ thoughts or ideas. \emph{Task-related noise} is correlated with the task execution and originates from the undesired activity of the subject, including irrelevant motion or cognitive processes performed during task execution.

\subsection{The need for noise identification in fMRI studies}

  As the interest to resting-state fMRI grows, there is an increasing number of noise studies related to task-free paradigms. Resting-state fMRI data contain coherent fluctuations unrelated to neural processes originating from residual motion artifacts, respiration and cardiac action~\cite{birn2012role}. These fluctuations may introduce spurious correlations and cause an overestimation of functional connectivity strength. Several multidimensional linear regression approaches to remove artificial coherencies were applied and the impact of different types of preprocessing on sensitivity and specificity of functional connectivity results in simulated data and experimental resting-state data was examined~\cite{weissenbacher2009correlations}. The authors tried to find possible causes of anticorrelations and find out whether these anticorrelations could be a result of certain preprocessing approaches like the regression against the global signal. Preprocessing, in general, was shown to greatly increase connection specificity; in particular, correction for global signal fluctuations almost doubled connection specificity~\cite{weissenbacher2009correlations}. Another finding was that widespread anticorrelated networks were only found when regression against the global signal was applied. Results both in simulated and human data strongly imply that anticorrelations could arise from global signal regression and, therefore, should be interpreted very carefully~\cite{weissenbacher2009correlations}. It was also shown that global signal regression could also reduce the sensitivity for detecting true positive correlations increasing the number of false negatives. Anyway, corrections against realignment parameters, white matter, and ventricular time courses, as well as the global signal are strongly recommended to maximize the specificity of positive resting-state correlations~\cite{weissenbacher2009correlations}. Presently, no established agreement on data cleaning exists as there is not a single “right” way to process resting state data that reveals the “true” nature of the brain~\cite{murphy2017towards}. The authors insist that further work is needed, different processing approaches are likely to reveal complementary insights about the brain's functional organization~\cite{murphy2017towards}. The doubts are strengthened by a recent work~\cite{bright2017potential}, where the authors describe potential pitfalls of denoising resting-state fMRI data using nuisance regression. Statistical assumptions of the GLM after nuisance regression of resting state fMRI data were examined. It was shown how pre-whitening, temporal filtering and temporal shifting of regressors could impact model fit~\cite{bright2017potential}. The authors make some recommendations: pre-whitening should be applied to achieve valid statistical inference of the noise model fit parameters as well as temporal filtering should be incorporated into the noise model to best account for changes in degrees of freedom~\cite{bright2017potential}.
  
    Aforementioned studies mainly analyze conventional (usually mass-univariate) statistical inference, which is not always the case when machine learning techniques are applied to fMRI data. One important task in resting-state fMRI is to find and compare similar resting-state networks (RSNs) across subjects.  Different machine learning algorithms were applied to tackle this problem~\cite{de2007classification}. SVM was shown to be successful in RSN classification~\cite{wang2015dimensionality} obtained by independent component analysis (ICA).  However, when applied to resting state fMRI, where many ICs reflect artifacts in data, classification of spatial components may generate an unreliable result, due to the major of complicated parameter tuning and overfitting~\cite{ren20173}.

\cite{mandelkow2017effects} provide evidence against the utilization of higher spatial resolution in stimuli decoding MVPA experiments due to noise issues. Machine learning approaches such as linear discriminant analysis (LDA) reach a maximum in classification performance at a smoothed resolution close to 3 mm, much above the 1.2 mm voxel size of the 7T fMRI acquisition. PCA of the global and local fMRI signal patterns suggest that informative neuronal fluctuations were spatially correlated and smooth, while other components of higher spatial frequency were more often related to physiological noise and responsible for the reduced classification accuracy at higher resolution \cite{mandelkow2017effects}.

  In a recent work, MVPA for each voxel was shown to not only make a distinct contribution to the information represented collectively by the population of voxels but to be highly sensitive to noise correlations between voxels~\cite{bejjanki2017noise}. The authors demonstrated that accurate decoding of the information content of neural populations these noise correlations should be taken into account. As shown before~\cite{da2014high}, multivariate decoding is enhanced for heterogeneous neural populations with high noise correlations. Based on this,~\cite{bejjanki2017noise} showed that noise correlations between heterogeneous populations of voxels influence MVPA. The authors repeated the analysis across different magnitudes of noise correlations and showed that it was positively related to classification accuracy. MVPA was shown to assign greater weights to voxels with high noise correlations. With artificial fMRI data, the authors were able to simulate the complementary effects of noise correlations and selectivity on decoding and show that these results remain the same for different parameter values~\cite{bejjanki2017noise}.

\subsection{Exploiting noise-related facts: fMRI simulation, noise identification and elimination techniques}

\subsubsection{fMRI data simulation}

  One of the two main purposes to simulate data is to optimize actual (real world) experiment design, while the other is to verify and compare statistical models/methods utilizing the fact that ground truth is known. The second point represents great relevance for the noise reduction as it means that information about the types and origins of noises is incorporated into the data simulation process, allowing the researcher to evaluate proposed classification and pattern recognition approaches. Currently, there is one major limitation in data generation - there is no well-established approach to model resting-state data, which is essentially presented as random fluctuations or (permuted) blocks of real background fMRI signal.

\subsubsection{Noise identification and suppression}

  Three significantly different general approaches for noise identification and removal in fMRI data can be highlighted~\cite{salimi2014automatic,caballero2017methods}: the first is based on using additional sensors measuring physiological activities exploitation (model-based approach,~\cite{glover2000image}), the second is noise elimination specific for each type of noise (e.g. motion correction or thermal noise cleaning), finally the third one is data-driven using only fMRI data itself and prior information about fMRI signal and noise. The first approach is limited as it covers only physiological nature of the noise and can’t handle e.g. scanner artifacts. Moreover, large amounts of data already been collected (and are being collected) without additional “noise” information, so the aforementioned method cannot be useful here. Data collection with additional equipment introduces additional challenges from increased experimental time to equipment cost, MR-compatibility, and instability. 

  Independent Component Analysis (ICA) based technique is could be viewed as a second step as it might be applied to components extracted by PCA~\cite{rasmussen2012nonlinear}. The resulting independent components are assumed to be either noise-related or signal-related, each representing one of the sources in a source separation task solved by ICA. ICA transforms source fMRI signal into a set of components with distinct spatial and temporal structures, which further could be classified as noise or signal. Three possible approaches to this classification can be highlighted. 

  The first one is an expert-based technique, meaning that an individual with expertise in fMRI processing must examine each component (its time courses, spatial distribution, and spectrum) and manually label it as either signal or noise~\cite{griffanti2017hand,kelly2010visual}. \cite{griffanti2017hand} present a detailed guideline for evaluating and categorizing independent components and provide examples of each component class. Expert-based categorization may be tedious and error-prone for data with low SNR ratios such as the case for ubiquitous medical 1.5 T scanners. In addition, this approach is extremely time-consuming as the number of independent components for each subject is normally around 50, and their visual inspection involves examination of a multitude of sources, including their spatial maps in three different projections, time courses of the respective independent component, and its frequency spectrum. While data quality represents an issue for all reviewed approaches, we note that manual examination is expected to provide the highest quality labeling when confronted with previously unseen data of low SNR.

  Another option is to utilize a pre-trained classifier such as the one provided by the FIX package of FSL that achieves 99\% classification accuracy based on the annotation created by human experts~\cite{salimi2014automatic}. In this work, ICA components are used to extract features for the machine learning methods (supervised learning classifiers) that aim to classify noise components from signal automatically based on labeled training data. 46 temporal and 131 spatial features are extracted and a feature selection procedure is performed during classifier training. This approach requires significantly less effort compared to an expert-based annotation but might result on lower quality noise suppression due to differences in experimental design and scanner parameters. 

  Finally, the third approach is to combine the first two approaches, i.e. to calibrate the existing model pre-trained on vast amounts of data with different characteristics (such as the one provided by the FIX package) for the particular problem. This requires creating a new task-oriented labeled dataset using the expert knowledge and using transfer learning techniques known from machine learning to 'fine-tune' the classification model on the newly labeled data. This approach seems to be the most promising when data quality is low and number of patients is relatively high. 

  Described approaches to signal-noise separation for fMRI data might prove useful for classification tasks in the medical domains described above, specifically epilepsy, schizophrenia, and depression diagnostics/prognosis. The crucial point here is that physiological noise having no direct relation to the neuronal activity (i.e. signal), might still carry valuable discriminating information for the classification task. For instance, cerebral blood flow fluctuations might reveal unobservable brain states, which correlate with the target variable (disease/no disease). Broadly speaking, two ways to approaching the classification problem exist. The first assumes building classifier based on the hand-crafted features extracted from the independent components (such as, for instance, described in~\cite{salimi2014automatic}), that could prove effective for discriminating between patients vs. healthy controls. An alternative approach may be based on the reconstruction of the 4D fMRI signal itself after noise elimination and its utilization as a source data for training (i.e. data might be denoised, or its signal and noise parts might be investigated separately).

\section{Advances in graph-based modeling of functional connectivity}

\subsection{Causal or effective connectivity and its applications to pattern recognition}

    All approaches discussed earlier mostly refer to the concept of functional or correlational connectivity, where the information of data (spatial) correlation structure is preserved. A relatively new neuroimaging concept is the effective (causal) connectivity (EC), which is defined as “influence that one neural system exerts over another, either at a synaptic or population level”~\cite{friston2011functional}, which means causation and, if estimated, would show the essence of neuronal processes of information transfer and processing in the brain. Effective connectivity is becoming very popular in fundamental neuroscience. Methods to assess EC are being developed nowadays, and they are split into two main groups: model-based like Dynamic Causal Modelling (DCM)~\cite{friston2011functional} and model-free, which are mainly based on information theory and its measures like conditional mutual information, Transfer Entropy (TE)~\cite{schreiber2000measuring,vicente2011transfer} and momentary information transfer~\cite{runge2014detecting}. Though applications of aforementioned methods to fMRI data could be rather tricky due to its dimensionality and autocorrelation structure, there is a growing number of (mostly fundamental) studies where DCM~\cite{dima2011effective,razi2015construct,sharaev2016causal} or TE~\cite{sharaev2016effective,sharaev2018information}. 
  
    Because of estimation complexity an (possible) instability in EC, up to now there are only a few examples of EC features use in classification and prediction tasks. One of them is dedicated to autism signatures~\cite{deshpande2013identification} and is focused on including EC measures in the predictive model. A multivariate autoregressive model (MVAR) was used to obtain the causality matrices for each of the 30 participants. Causal connectivity weights were added to feature space of functional connectivity values and passed to support vector machine classifier to determine the accuracy of autism versus control discrimination. Maximum classification accuracy was found to be 95.9\% with 19 features with the highest discriminative ability between the groups. All of the 19 features were effective connectivity graph paths, indicating that causal information may be critical in discriminating between autism and control groups~\cite{deshpande2013identification}. 
  
    Another study aimed at classification between major depressive disorder (MDD) patients and healthy controls~\cite{geng2018multivariate}. The authors used the whole brain connectivity measures which included the ROI-to-ROI functional connectivity from Automated Anatomical Labeling (AAL) template and EC between resting-state nodes, such as the default mode network (DMN), dorsal attention network (DAN), frontal-parietal network (FPN), and silence network (SN). Effective connectivity was assessed using spectral Dynamic Causal Modeling (spDCM). Linear SVM, non-linear SVM, KNN  and Logistic Regression (LR) were tested as classifiers to discriminate between MDD patients and healthy controls. The highest accuracy was 91.67\% (p < 0.0001) when using 19 effective connections and 89.36\% when using 6,650 functional connections. The authors also performed a classification analysis using only functional connections from aforementioned resting-state networks where the highest accuracy was 78.33\% (p < 0.0001). So, despite the fact that regions of interest were the same, EC provided higher accuracy compared to FC~\cite{geng2018multivariate}.

\subsection{Topological data analysis}

  Recent studies have demonstrated the successful application of topological approaches for connectome construction and analysis~\cite{dlotko2016topological}. As traditional graph theory approaches may be insufficient to understand the full extent of complexity of such an intertwined object as the brain network (or even a small area of it), the work has investigated algebraic topology approaches to define its structural and functional organisation. Structural topological data analysis has demonstrated the significant difference between micro-level connectivity networks of the inferred graph vs. various kinds of random graphs. In particular, oriented graphs contain on the order of $10^7$  fully-connected simplex groups. Some of these simplexes contain up to 8 neurons, making them the largest fully-connected motif ever discovered. Functional topological analysis of simulated neuronal activity data has revealed novel spatio-temporal measures providing efficient classification of functional responses to qualitatively distinct kinds of stimuli. The methods proposed in this work demonstrate general applicability in connectomics and neuroscience.
    
  Other study uses topological data analysis to investigate the functional networks obtained from both experimental and simulated time series~\cite{stolz2017persistent}. Coupled Kuramoto oscillators serve as a source of simulated time-series data, while the experimental data relate to the fMRI recording of healthy subjects made when subjects performed a simple motor task. The experiments have demonstrated that the persistent homology is a promising new tool for the analysis of functional networks. Persistent homology-based analysis has been demonstrated to reveal the differences in synchronisation patterns in the datasets over time, while simultaneously emphasizing the changes in the cluster structure of the graph network. It also reveals the increase in the degree of synchronisation between brain regions leading to the emergence of brain circuits while learning to perform a motor task. The method has helped to establish that the most significant changes in brain circuits emerge during the second day of study.

\section{Noise-aware fMRI processing pipeline for~pattern recognition with~applications to~neurology}

\subsection{The proposed pipeline}

\begin{figure}
\centering
\includegraphics[width=0.8\textwidth]{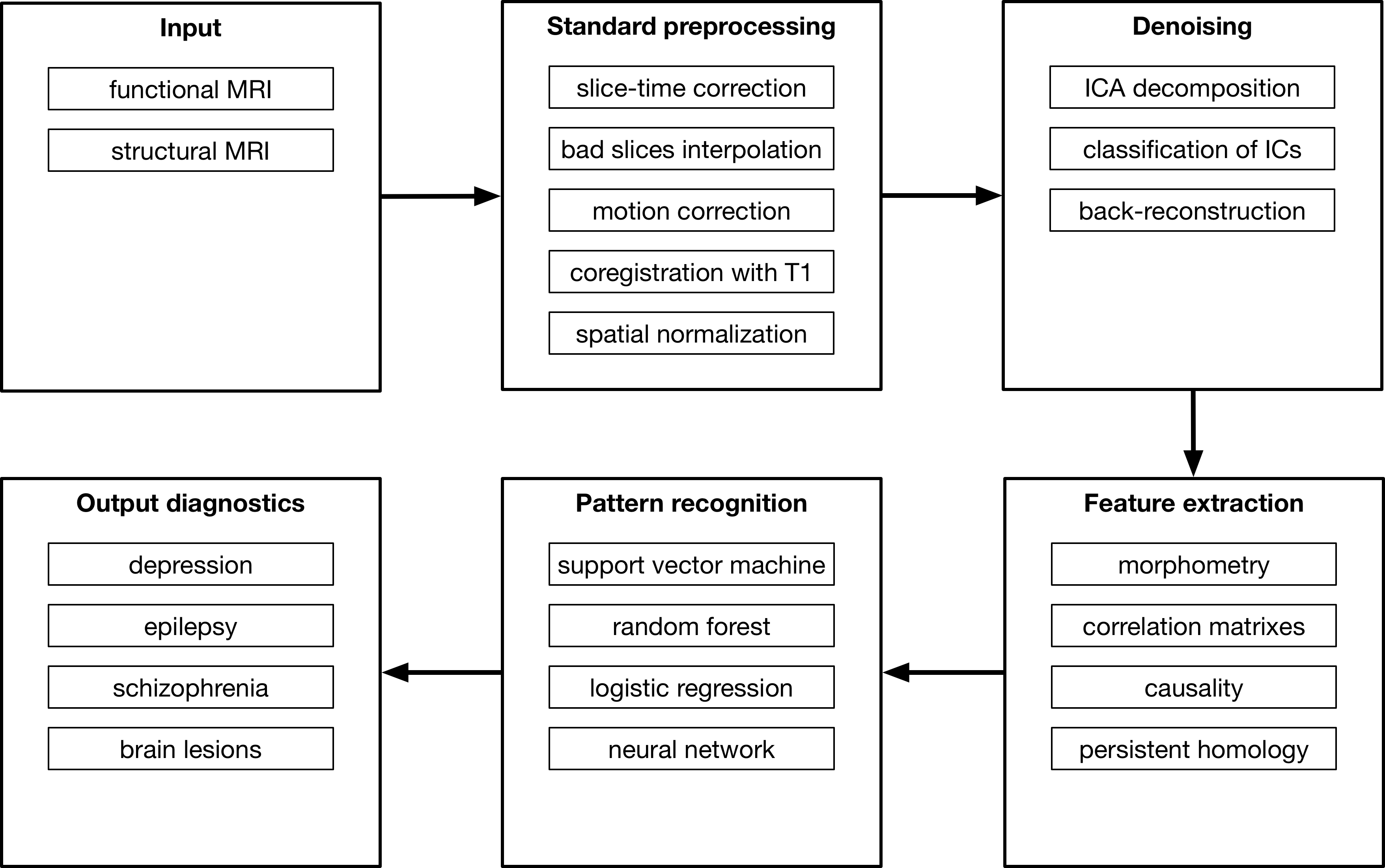}
\caption{\label{fig:scheme}The scheme of the proposed noise-aware fMRI processing pipeline.}
\end{figure}
  
    The literature review has allowed us to identify established and prospective building blocks and organize them into a unified and highly automated fMRI processing pipeline. Our pipeline accepts raw functional and structural scans of a subject and outputs the predicted task-specific scores, whose meaning vary according to the application. For instance, for a depression vs. healthy control classification task, our pipeline should score the patient according to a probability of depression diagnosis for him. The entire chain of steps can be implemented in a modular way via existing or prospective software by respecting interfaces between the modules. We note that as some of the modules may carry computationally intensive processing, the runtime of the pipeline may vary from minutes to several hours. We briefly describe our proposed pipeline below.

  The input to our pipeline comprises of functional and structural MRI scans. The raw scans are passed through a standard preprocessing step, an established low-level MRI handling stage (involving slice-time correction, motion correction, filtering in space and time, anatomical alignment across individuals)~\cite{cohen2017computational}, yielding preprocessed data on the same scale and format. A second stage accepts preprocessed scans and runs a denoising procedure analogous to the one implemented in FIX software~\cite{salimi2014automatic}, producing a scan with irrelevant components excluded and an increased SNR. We note that, in fact, the independent components marked as “noise” may still carry discriminative features for the end classification task, thus, it may be beneficial to keep those instead of excluding them from the consideration altogether. The third stage of our pipeline performs morphometric, correlation-based, causal, and graph-based feature extraction. We emphasize the perspective use of graph-based and topological features for the pattern classification problem, as these have been long indicated as relevant for pattern classification problem~\cite{wang2017depression,stolz2017persistent}. Lastly, our pipeline performs pattern recognition by making use of available implementations of conventional machine learning approaches such as SVMs, neural networks, and decision trees, to name a few.
  A schematic overview of our proposed pipeline can be seen in Figure~\ref{fig:scheme}.

\subsection{A pilot study}
  The purpose of this example study is to demonstrate the possible advantage of using novel sophisticated artifact removal procedures in clinical diagnostics. The task was to distinguish between two groups of patients: only epilepsy and both epilepsy+depression patients. Whereas structural MRI data could be enough to distinguish between epilepsy and healthy controls, the more precise diagnostics as our task was, needs to take fMRI into consideration. Our data was 25 patients with epilepsy and 25 both epilepsy+depression patients. Collected at 1.5T EXCEL ART VantageAtlas-X Toshiba scanner by physicians from Russian Scientific and practical psycho-neurological center named after Z.P. Solovyov (NPCPN http://npcpn.ru/) (Skoltech biomedical partner).
  
    Raw data was preprocessed in three different ways: 1) “Standard” - standard preprocessing pipeline in SPM12 including slice-timing correction, bad-slices interpolation, motion-correction (bad volumes interpolation), coregistration with T1 images and spatial normalization; 2) “Standard+FSL FIX denoise” is standard pipeline plus removing noise ICA based on FIX classification (FIX was not trained specifically on our data, but pre-trained by developers on high-quality 3T fMRI data) and 3) “Standard+manual denoise” which is a combination of standard pipeline with manual ICA classification into signal and noise components by fMRI experts. For the parcellation of the brain, an Automatic Anatomic Labeling (AAL) atlas was used, which consist of 117 regions. For each region corresponding time series were assigned, and then a correlation matrix was calculated from them.
  
    To work with graphs, we used the Python library - networkX 2.1. The adjacency matrix for the graph was obtained from the correlation matrix, where the values greater than the threshold were replaced by 1, and the others by 0. We calculated 5 metrics corresponding to each region of the brain: clustering coefficient, degree centrality, closeness centrality, betweenness centrality, average neighbor degree and 2 metrics witch describe the graph in general: local efficiency and global efficiency. As a result we obtain a matrix with $5 \times 117 + 2 = 587$ features. Then we use standard machine learning classifiers: Support Vector Machine (SVM), Random Forest Classifier (RF), and Logistic Regression (LR).
  
    Each model was validated using leave-one-out approach which gave the following results:
(1) “Standard” with accuracy of $0.72 \pm 0.13$,
(2) “Standard + FSL FIX denoise” with accuracy of $0.52 \pm 0.14$, and 
(3) “Standard + manual denoise” with accuracy of $0.88 \pm 0.09$.
Table~\ref{table:denoise-quality-tpr-fpr} provides details on the quality of our approach.

\begin{table}[]
\centering
\caption{True Positive Rates for three assessed methods with fixed False Positive Rate.}
\label{table:denoise-quality-tpr-fpr}
\begin{tabular}{llll}
\multirow{2}{*}{\textbf{False Positive Rate}} & \multicolumn{3}{c}{\textbf{True Positive Rate}}                            \\
                                     & Standard & Standard + manual denoise & Standard + FSL FIX denoise \\
                                     \hline
0.1                                  & 0.2      & 0.76                      & 0.02                       \\
0.15                                 & 0.28     & 0.92                      & 0.08                       \\
0.2                                  & 0.36     & 0.92                      & 0.08                       \\
0.3                                  & 0.44     & 0.92                      & 0.08                      
\end{tabular}
\end{table}

  Firstly, it can be seen that “Standard” preprocessing does not perform well when working on simple features (functional connectivity). “Standard+FSL FIX denoise” performance is even worse implying that additional training of FIX package is almost obligatory on different datasets (most likely due to different imaging quality of 1.5T and 3T MR-scanners). Finally, “Standard+manual denoise” shows relatively high performance in terms of accuracy and true positive rate, which means that additional sophisticated data cleaning could be beneficial for fMRI classification tasks.

\section{Conclusions}

  Functional neuroimaging has proven promising for clinical diagnostics of psychiatric disorders such as depression and neurological diseases such as epilepsy.  In current work we reviewed some common and established approaches to preparation, preprocessing and denoising of neuroimaging data aimed at the elimination of artifacts harmful for pattern recognition. We also reviewed modern advanced approaches to data representation for pattern recognition, graph-based methods and topological data analysis as the promising areas. Based on well established and modern approaches, we proposed a principled noise-aware pattern recognition pipeline for neuroimaging tailored to pattern classification and showed the potential effectiveness of our proposed methodology in a pilot clinical rs-fMRI classification study. As a result denoised data provides clearer and more informative features for machine learning-based diagnostics, and yields significant improvements in classification accuracy between a group of patients with only epilepsy versus epilepsy+depression patients. 

\acks\begingroup
This study was performed in the scope of the Project ``Machine Learning and Pattern Recognition for the development of diagnostic and clinical prognostic prediction tools in psychiatry, borderline mental disorders, and neurology'' (a part of the Skoltech Biomedical Initiative program). %with collaboration of physicians from Russian Scientific and practical psycho-neurological center named after Z.P. Solovyov (NPCPN http://npcpn.ru/) (Skoltech biomedical partner), which provided medical data and biomedical expertise. 
\endgroup

\bibliography{references/references.bib}

\end{document}